\pdfoutput=1
\documentclass[10pt]{article}

\usepackage[top=4cm,bottom=3cm,left=2.5cm,right=2.5cm]{geometry}
\usepackage[numbers,sort&compress,sectionbib]{natbib}

\usepackage{subcaption}
\usepackage{graphicx}

\usepackage{tkz-tab}
\usepackage{amsmath}
\usepackage{graphicx}
\usepackage[colorlinks=true, allcolors=blue]{hyperref}
\usepackage{graphicx} 
\usepackage{amssymb} 
\usepackage{hyperref}
\usepackage{enumerate}
\usepackage{neuralnetwork}
\usepackage{cancel}
\usepackage{multirow}
\usepackage{float}
\usepackage{bm}
\usepackage{xspace}
\usepackage{tikz,pgf}
\usepackage{pgfplots}
\usepackage{graphicx,color,stackrel,xcolor}
\usepackage{array}
\usepackage{everysel}
\usepackage{makecell}
\usepackage{multirow}
\usepackage{ragged2e}
\usepackage{wrapfig}
\usepackage{booktabs} 
\usepackage{soul}
\newcommand{\comment}[1]{}
\newcommand{\iouibp}{IBP IoU\xspace}

\usepackage{adjustbox} 
 \pgfplotsset{compat=1.13}

\newtheorem{hypothesis}{Hypothesis}
\newtheorem{definition}{Definition}
\newtheorem{theorem}{Theorem}
\newtheorem{property}{Property}

\newcommand{\interval}[1]{\ensuremath{\left[\underline{#1}, \overline{#1}\right]}}
\newcolumntype{M}[1]{>{\centering\arraybackslash}m{#1}}
\newcommand{\iou}{\textit{IoU}\xspace}
\newcommand{\iougt}{\ensuremath{IoU_{gt}}\xspace}

\newcommand{\dlocmaj}{DIGIT\_LOC}
\newcommand{\bbox}{bounding box\xspace}
\newcommand{\bboxes}{bounding boxes\xspace}
\newcommand{\fig}{Figure}

\newcommand{\symbolbbox}{b}
\newcommand{\symbolbboxzero}{\ensuremath{b_0}}
\newcommand{\symbolbboxone}{\ensuremath{b_1}}
\newcommand{\areabo}{\ensuremath{a(b_0)}}
\newcommand{\areabone}{\ensuremath{a(b_1)}}
\newcommand{\areabobone}{\ensuremath{a(i_{b_0}(b_1))} }

\makeatletter
\renewcommand\subsubsection{\@startsection{subsubsection}{3}{\z@}%
                       {-18\p@ \@plus -4\p@ \@minus -4\p@}%
                       {0.5em \@plus 0.22em \@minus 0.1em}%
                       {\normalfont\normalsize\bfseries\boldmath}}
\makeatother
\title{VerifIoU - Robustness of Object Detection to Perturbations}
\author{Noémie Cohen$^1$, Mélanie Ducoffe$^1$, Ryma Boumazouza$^1$, Christophe Gabreau$^1$,\\
  Claire Pagetti$^2$, Xavier Pucel$^2$ and Audrey Galametz$^1$\\
$^1$ Airbus, $^2$ French Aerospace Lab, ONERA}

\begin{document}

\maketitle

\begin{abstract}
  This paper addresses the challenge of verifying the robustness of object detection models in safety-critical applications, such as aeronautics. Focusing on vision-based aircraft pose estimation, the study aims to ensure that perturbations do not degrade the model's ability to accurately localize runways. A key challenge arises from the Intersection over Union (IoU) metric used in object detection, which complicates formal verification due to its non-convex and multidimensional nature. 
We propose a method, IBP-IoU, to improve precision, computational efficiency and completeness in verification.
The method bridge the gap between classification and object detection verification and are demonstrated through aeronautical and digit localization case studies enabling verification for single object detection.

\end{abstract}

\section{Introduction} \label{sec:introduction}

The emergence of Machine Learning (ML) and, in particular, deep learning and neural network (NN) models has allowed new capabilities for a wide range of application domains e.g.,~transportation, healthcare, finance etc. However, ML techniques often show intriguing properties. An extensive literature has shown NN vulnerabilities to adversarial examples e.g.,~\cite{szegedy2013intriguing}. This inherent flaw in neural networks presents a significant challenge for the development of ML-based safety-critical applications. It is therefore essential to explore tools that provide formal correctness guarantees to ensure ML \emph{robustness} and prevent potential safety risks.

Historically, formal verification methods have already been used by AIRBUS in traditional development. 
The use of formal methods is motivated by the expectation that performing appropriate mathematical analyses can contribute to establishing the correctness and robustness of a design.
A supplement, ED-216/DO-333 \cite{do333}, is available for employing formal methods as a means of providing evidence that verification objectives are met. Recent regulatory developments, such as the EU AI Act (May 2024), emphasize the need for technical robustness and safety, mandating that AI systems must be resilient to tampering and capable of minimizing unintended harm. In the aviation sector, the European Aviation Safety Agency’s (EASA) Artificial Intelligence (AI) roadmap and Concept paper (March 2024) explicitly highlight the necessity of preserving critical model properties, where the use of formal verification methods is a means to ensure compliance. Similarly, the Federal Aviation Administration (FAA) Roadmap for AI Safety Assurance (July 2024) underscores the need for new assurance methods, advocating for the establishment of criteria to select appropriate formal methods and testing tools. These regulatory trends highlight the urgent need for advanced verification tools for neural networks that provide a means to guarantee that ML models meet the safety requirements necessary for deployment in high-stakes environments.

Most of the published works on NN formal verification have focused on object classification tasks and addressed the scalability challenges of providing formal robustness guarantees for deep neural networks (DNNs) (e.g., \cite{katz2017reluplex}). The present work is motivated by the challenge to extend these verification works to object detection models, especially with regards to their increasing use in industries such as autonomous driving for real-time obstacle detection. An object detection model is a machine learning system designed to identify and locate objects within images or video frames by providing bounding boxes and class labels for each detected object.

In this paper, we introduce an approach to formally assess the robustness of object detection models against local perturbations. The accuracy of such models is commonly evaluated using the Intersection over Union (IoU) which represents the match between the actual location of the object on the image (ground truth) and the model prediction. 

\begin{figure}[hbt]
\centering
\captionsetup[subfigure]{justification=centering}    
\begin{subfigure}{0.24\linewidth}
    \centering
        \includegraphics[width=\linewidth]{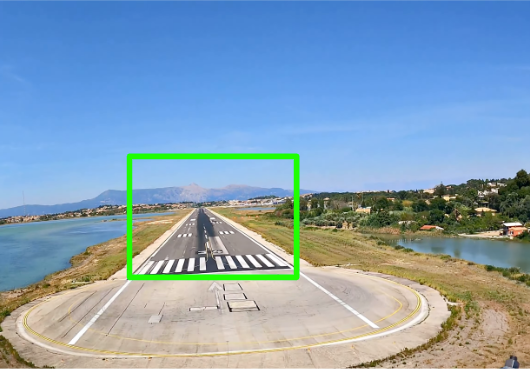}
        \caption{Reference input and ground truth}
        \label{fig:impact:ok}
    \end{subfigure} %
    \begin{subfigure}{0.24\linewidth}    
    \centering
        \includegraphics[width=\linewidth]{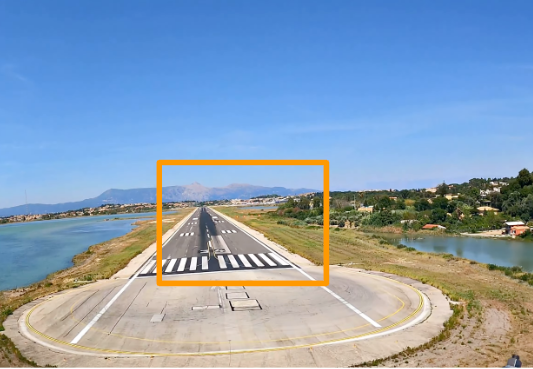}
        \caption{Slight impact,\\ IoU=$0.65$}
        \label{fig:impact:slight}    
    \end{subfigure} 
     \begin{subfigure}{0.24\linewidth}
     \centering
        \includegraphics[width=\linewidth]{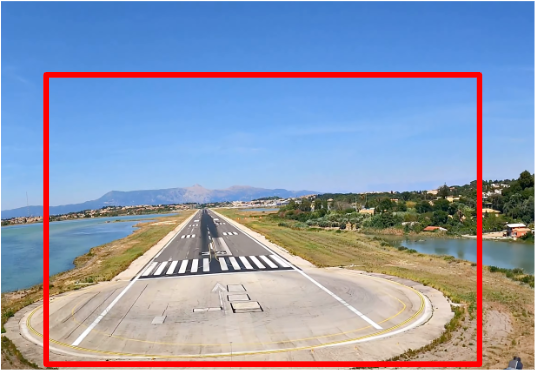}
        \caption{Strong impact, IoU=$0.17$}
        \label{fig:impact:strong1}
    \end{subfigure}
    \begin{subfigure}{0.24\linewidth}
     \centering
        \includegraphics[width=\linewidth]{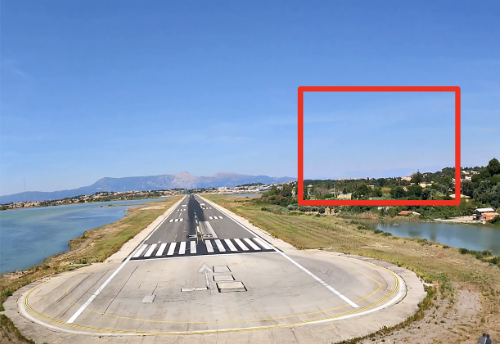}
        \caption{Strong impact, \\ IoU=$0$}
        \label{fig:impact:strong2}
    \end{subfigure}
    \caption{
    Impact of input perturbations 
    captured by the IoU.
    }
    \label{fig:impact_of_perturbation}
\end{figure}

We focus on bounding the extreme values of the IoU, a challenge due to its multi-dimensional, non-convex nature. \\

The main technical contributions of the paper are as follows.
\begin{itemize}
    \item Addressing the significant gap in existing verification methods, which have primarily focused on classification tasks.
    \item Introducing a solver-agnostic approach, allowing compatibility with various solvers. 
    \item Establishing IoU as a key component for verification; in addition to the technical contributions, this paper lays the foundation for verifying object detection models.
    \item Demonstrating the approach on an industrial use case, in addition to standard academic datasets. 
\end{itemize}

\section{Related Work} \label{sec:relatedwork}

Ensuring the reliability of object detection models through formal verification has emerged as a crucial challenge, especially with regards to safety (e.g., \cite{kouvaros2023verification}). 

{\bf Empirical approaches: }Adversarial attacks are carefully crafted perturbations to input data that fool a model into making incorrect predictions \cite{szegedy2014intriguing}. These attacks can take various forms, including targeted and untargeted attacks, where the goal is either to force a model to misclassify a specific input or to cause a general deterioration in the model performance. Different techniques navigating on the trade-off between computing time and fooling rate have emerged for classification task such as \cite{carlini2017evaluating}, \cite{moosavidezfooli2016deepfool} and \cite{abdollahpourrostam2023revisiting}. More recently, the landscape of adversarial attacks has expanded to include generalized attacks designed specifically for Object Detection (OD) tasks. \cite{chow2020tog} in their work on Targeted Adversarial Objectness Gradient Attacks
on Real-time Object Detection Systems (TOG) introduced three targeted adversarial Objectness Gradient attacks that exploit specific vulnerabilities in object detection systems, such as making objects vanish (attack causes all objects to vanish), fabricating false objects (output many false objects with high confidence), and mislabeling objects (attack fools the detector to mislabel). The authors also develop a highly efficient universal adversarial perturbation algorithm, capable of fooling object detectors in real-time with minimal online attack cost, posing a significant threat to real-time edge applications. Despite the progress in understanding and categorizing adversarial attacks, existing defenses often fall short in providing strong guarantees of model robustness. Most current approaches rely on empirical testing, which, while useful, does not offer comprehensive assurances against all possible adversarial scenarios. This gap highlights the need for formal methods that can provide rigorous, mathematically proven guarantees of robustness.

{\bf Formal methods:} These methods include exact approaches, where all possible model behaviors are exhaustively analyzed, and abstraction-based methods, where the model's behavior is approximated using techniques like convex relaxation. Authors in \cite{bastani2016measuring} expressed the robustness problem as a satisfiability checking of a logical formula encoding the NN semantics and the properties. For instance, encoding formulas into a linear real arithmetic enables the use of verifiers based on the Satisfiability Modulo Theory (SMT) (e.g., \cite{katz2017reluplex,ehlers2017formal,huang2017safety,tjeng2017evaluating}) and Mixed Integer Linear Programming (MILP) solvers (e.g., \cite{bastani2016measuring,tjeng2017evaluating,katz2019marabou,botoeva2020efficient}). It is worth noticing however that current formal verification methods still suffer from scalability issues along with a limited number of types of perturbations and NN layers currently supported. \\
On the other hand, abstract interpretation allows for tractable verification by simplifying the complex, non-linear decision boundaries of neural networks into linear or convex forms that can be more easily analyzed \cite{urban2021review}. The current state-of-the-art in formal verification of neural networks is dominated by linear relaxation methods, LiRPA, for Linear Relaxation based Perturbation Analysis. The LiRPA techniques have been further explored as demonstrated by tools such as ERAN \cite{eranpaper}, Auto-LIRPA \cite{autolirpa} or DECOMON \cite{decomon}. These techniques were applied in the International Competition on Verification of Neural Networks (VNN-Comp). 

{\bf Formal methods for object detection:}  IoU is commonly used in computer vision to evaluate object detection models because it directly measures the overlap between predicted and ground-truth bounding boxes, providing a clear assessment of localization accuracy. This simplicity and effectiveness make it a standard metric in the field. In 2023, the VNN-Comp included a dedicated section of benchmarks focussing on object detection challenges. None of these tasks consider the robustness of the object detection localization, measured with the IoU. For example, the benchmark of \cite{collins_benchmarks_vnn_comp} focuses on the robustness of the objectness score, defined as the confidence that a given region in an image contains an object of interest, regardless of its specific class. 
The \cite{traffic_sign_benchmarks_vnn_comp} benchmark did consider the robustness of IoU, but only under a limited set of perturbations that did not require competitors to adapt any existing solvers. As a result, competitors only evaluated the IoU metric across perturbed samples. For the 2024 edition, a benchmark \cite{iou_segmentation_benchmarks_vnn_comp, iou_segmentation_original_paper} includes IoU but for segmentation tasks. Although the name is the same, the underlying function differs. Specifically, for segmentation tasks, the robustness of IoU is expressed as a piecewise linear function based on the output of the object detection model. Therefore, the robustness analysis for segmentation tasks is already compatible with existing solvers. 

The only work that has emerged later in this direction (formal verification for Object Detection) is by \cite{raviv2024formal}, where they encoded the IoU as a neural network using operators already supported by LiRPA solvers. This approach adapts existing formal methods to the unique challenges posed by metrics like IoU, which are critical for evaluating OD model performance. However, our approach differs fundamentally from that of \cite{raviv2024formal}. In their work, the IoU function is treated as a latent layer within the network, which simplifies the verification process. In contrast, we approach the IoU as a metric rather than a network layer, which requires a more nuanced analysis of its extreme (minimum and maximum) values. This 
metric-centered approach brings our work closer to the verification challenges seen in classification tasks, where metrics like cross-entropy are central. Similar to the work of \cite{huang2021training} on training verifiably robust models, we explore how these verification techniques can be extended and applied to the unique demands of object detection metrics.

\section{Common concepts in object detection}
\label{sec:robustnessODperturbations}

In the present paper, we consider models whose task is to perform the detection of one single object, ideally delineating it using a tight bounding box. We also consider one type of object/class. Let's (re)introduce some general concepts on object detection models.

\begin{definition}[Bounding box]
It is a rectangle that encapsulates the object of interest. We define a bounding box \symbolbbox~$= [z_0, z_1, z_2, z_3]$ with ($z_0$, $z_1$) and ($z_2, z_3$), the (x,y) coordinates of the bottom-left and upper-right corners of the box. We define the set of \bboxes as $\mathcal{B} = \{ [ z_0, z_1, z_2, z_3] \in \mathbb{R}^4_{+} \mid z_0\leq z_2, z_1\leq z_3\}$.
\end{definition}

The concept of bounding box can refer to a ground truth i.e., the box around the actual object. We will here refer to a ground truth bounding box as $b^{gt} = [z_0^{gt},z_1^{gt},z_2^{gt},z_3^{gt}]$. Bounding boxes can also refer to the model prediction. The model computes a set of candidate bounding boxes, and only returns the box with the highest `objectness' score refering to the box with the highest confidence. 

\begin{definition}[Single class/single object detection model] If $s_0$ an input image of dimension $n$, and \symbolbbox, the bounding box with the highest objectness score. A single class / single object detection model is a function $f_{OD}$ defined by: 
\begin{equation}
\label{equ:fod}
\begin{array}{lcll}
f_{OD}:\; &\mathcal{X} \subseteq \mathbb{R}^n & \longmapsto &\mathbb{R}^{4}\\
& s_0& \longrightarrow &\text{\symbolbbox}
\end{array}
\end{equation}
\end{definition}

Intersection over Union (\iou) 
\cite{rezatofighi2019generalized} is a metric 
that quantifies box overlap by calculating the ratio of their intersection area to their union area. 

\begin{definition}[Intersection over Union -- \iou]\label{def:iou} 
Let a reference \bbox \symbolbboxzero $=[z_0^0, z_1^0, z_2^0, z_3^0]$ and its area $a(b_0)$ defined by the function $a: \mathcal{B} \longmapsto \mathbb{R}_{+} $ where 
\begin{equation}a(b_0)= (z_2^0 - z_0^0)\times (z_3^0-z_1^0)\label{area_predicted}
\end{equation}
Let a \bbox \symbolbboxone $=[z_0^1, z_1^1, z_2^1, z_3^1]$ and its area $a(b_1)$ similarly defined.
Let the intersection of $b_1$ with $b_0$, $i_{b_0}(b_1)$ defined by the function 
$i: \mathcal{B}^2 \longmapsto \mathcal{B}$,
\begin{equation}
i_{b_0}(b_1)=( \max\limits_{j=0,1} z_0^j, \max\limits_{j=0,1} z_1^j, \min\limits_{j=0,1} z_2^j, \min\limits_{j=0,1} z_3^j \big)
\label{intersection}
\end{equation}
and its area \text{\areabobone}. The \iou is a function $\mathcal{B}^2 \longmapsto [0, 1]$ such that:
\begin{equation}\label{equ:iou}
\iou_{b_0}(b_1)= \frac{\text{\areabobone}}{\text{\areabo} + \text{\areabone} - \text{\areabobone} }
\end{equation}
\end{definition}

\begin{figure}[hbt]
\centering
\begin{minipage}{0.5\linewidth}
  \resizebox{.8\linewidth}{!}{
    \begin{tikzpicture}[thick,scale=0.4, every node/.style={scale=0.7}]
\draw[step=1cm,gray,very thin] (-1,-1) grid (9,8);
\draw[thick,->] (0,0) -- (7.5,0);
\draw[thick,->] (0,0) -- (0,6.5);
\draw[thick,->] (0,0) -- (7.5,0) node[anchor=north west] {x};
\draw[thick,->] (0,0) -- (0,6.5) node[anchor=south east] {y};
\foreach \x in {0,1,2,3,4,5,6,7}
   \draw (\x cm,1pt) -- (\x cm,-1pt) node[anchor=north] {$\x$};
\foreach \y in {0,1,2,3,4,5,6}
    \draw (1pt,\y cm) -- (-1pt,\y cm) node[anchor=east] {$\y$};
\fill[red!40!white] (4,5) rectangle (1,3);
\fill[yellow!40!white] (6,4) rectangle (3,1);
\fill[orange!40!white] (4,4) rectangle (3,3);
\node at (4,5) {x};
\node at (4,5.5) {$(z_2^1, z_3^1)$};
\node at (1,3) {x};
\node at (1,2.7) {$(z_0^1, z_1^1)$};
\node at (6,4) {x};
\node at (6,4.3) {$(z_2^0, z_3^0)$};
\node at  (3,1) {x};
\node at (3,0.7) {$(z_0^0, z_1^0)$};
\end{tikzpicture}}
   \end{minipage}
\begin{minipage}{0.45\linewidth}
If $b_0$ and $b_1$ the yellow (reference) and red bounding boxes respectively:
\begin{itemize}
\item Intersection (orange): $b_0 \cap b_1 = i_{b_0}(b_1)$. 
\item Areas:
$a(b_0)=9$, $a(b_1)=6$ and $a(i_{b_0}(b_1))=1$. 
\item $IoU_{b_0}(b_1)=\frac{1}{14}$.
\end{itemize}
\end{minipage}
\caption{Quantitative example of \iou}
\label{fig:iou_def}
\end{figure}

Figure~\ref{fig:iou_def} provides a quantitative example. \iou$=1$ is a perfect match while \iou$=0$ means that the boxes do not overlap. In the context of object detection, the \iou metric refers to the match between the bounding box of the ground truth and the one predicted. We define $i_{gt}(b)$, the intersection of a predicted  box with respect to the ground truth and $IoU_{gt}(b)$, the associated \iou. $IoU_{gt}(b) > 0.5-0.6$ is commonly considered a good detection score because, in standard object detection benchmarks, predictions with an \iou above this threshold are counted as correct detections.

\section{Robustness of object detection} 
\label{sec:method}
This section provides an overview of the proposed formal verification solution for object detection models via the assessment of the robustness to perturbations of the \iou metric. 
Let us first explain which perturbations are considered 
and introduce notations for representing bounds.

\subsection{Perturbations applied to object detector}
We intend to assess how perturbations impacting an input image can affect the performance of a detection model (via the assessment of its impact on the \iou).

\begin{definition}[Perturbation domain]
\label{def:perturbationdomain}
Given an input image $s_0$, a local perturbation domain $\Omega(s_0)$ encompasses all images computed by applying a certain perturbation to $s_0$.  In the case of white noise perturbation for instance, the perturbation domain is defined using the \( l_\infty \)-norm ball:
$\Omega(s_0) = \{ s \in \mathbb{R}^n \mid \|s - s_0\|_\infty \leq \epsilon \}$ where \( \epsilon \) controls the perturbation magnitude.
\end{definition}

Different types of image perturbations have been investigated in the literature including local ones such as white noise, brightness and contrast \cite{kouvaros2023verification}. 
We here assume that the perturbations have no impact on the position of targeted object and thus, the ground truth \bbox.

\comment{
A noise perturbation domain consists of all images potentially obtained by applying an additive value to each pixel independently. The value of noise is usually limited to a certain threshold ($\pm \epsilon$). Brightness/contrast perturbation domains consist of all images obtained by applying a uniform additive/multiplicative coefficient $\alpha_b$/$\alpha_c$, respectively. We define these three perturbation domains in \fig~\ref{tab:perturbation}. 

\begin{figure}[hbt]
\centering
\begin{tabular}{l|c|c|ccc|}
Perturbation &  Factor & $\Omega(s_o)$ \\
\hline
White noise & $\epsilon$ &$\{s \in \mathbb{R}^n \mid||s-s_0||_\infty \leq \epsilon \}$\\
Brightness & $\alpha_b$ &$\{s \in \mathbb{R}^n \mid s= s_0+\alpha_b \}$\\
Contrast & $\alpha_c$ &$\{s \in \mathbb{R}^n \mid s= s_0\times\alpha_c \}$\\
\end{tabular}
\caption{Definition of common perturbations}
\label{tab:perturbation}
\end{figure}
}

\subsection{Notations related to bounds}
Incomplete formal verification commonly uses the concept of bounds to refer to the values derived to over-approximate the model prediction domain. In order to ease the understanding of our proposed object detection verification approach, we first introduce a few notations with respect to bounds. 

Let's consider a scalar $s$ where $s\in [\underline{s}, \overline{s}]$. $\underline{s}$ and $\overline{s}$ are lower and upper bounds of $s$ such that $\underline{s} \leq s \leq \overline{s}$. We extend these notions of bounds beyond scalars. 

\comment{
\begin{definition}[Bounds for arrays] 
\label{def:interval}    
Let's consider an array $i = [i_0, \ldots, i_N]$. We define bounds for $i$ such that $i \in \interval{i}$. $\underline{i} = [\underline{i_0}, \ldots, \underline{i_N}]$ and $\overline{i} = [\overline{i_0}, \ldots, \overline{i_N}]$ where, for each $i_j$, ${\underline{i_j}} \mathbf{\leq} {i_j} \mathbf{\leq} \overline{i_j}$.
\end{definition}
}

\begin{definition}[Bounds for predicted bounding boxes]\label{bbb} Let's define $\mathbf{b}$, the set of bounding boxes that can be predicted from a perturbation domain $\Omega(s_0)$ (where $s_0$, the original perturbed image) i.e., $\mathbf{b} = f_{OD}(\Omega(s_0))$. We define bounds for $\mathbf{b}$ such that $\mathbf{b} \in \mathbf{\interval{b}} = (\interval{z_0}, \interval{z_1}, \interval{z_2}, \interval{z_3})$. By design, $\underline{z_0} \leq \underline{z_2}$, $\overline{z_0} \leq \overline{z_2}$, $\underline{z_1} \leq \underline{z_3}$, and $\overline{z_1} \leq \overline{z_3}$.
\comment{$\mathbf{\underline{b}}\in \mathcal{B}$ and $\mathbf{\overline{b}}\in \mathcal{B}$ are bounding boxes.
The \emph{outer bound} of an extended bounding box $\mathbf{b}$ is the bounding box $out(\mathbf{b}) = [\underline{z_0}, \underline{z_1}, \overline{z_2}, \overline{z_3}]$. The \emph{inner bound} of an extended bounding box $\mathbf{b}$ is the bounding box $in(\mathbf{b}) = [\overline{z_0}, \overline{z_1}, \underline{z_2}, \underline{z_3}]$ if $\overline{z_0} < \underline{z_2}$ and $\overline{z_1} < \underline{z_3}$, otherwise it is the empty set.}
\end{definition}

\begin{definition}[Bounds for \iou] Given a set of bounds for predicted bounding boxes $[\underline{\mathbf{b}}, \overline{\mathbf{b}}]$, $IoU_{gt}(\mathbf{b})$ is the set of \iou that can be derived from $\mathbf{b}$ with respect to a ground truth bounding box $b_{gt}$. We define bounds for $IoU_{gt}(\mathbf{b})$ such that $IoU_{gt}(\mathbf{b}) \in [\underline{IoU_{gt}}(\mathbf{b}), \overline{IoU_{gt}}(\mathbf{b})]$.
\begin{equation}
    \forall\; \mathbf{b} \in \mathbf{\interval{b}} \implies \underline{IoU_{gt}}(\mathbf{b}) \leq IoU_{gt}(b) \leq \overline{IoU_{gt}}(\mathbf{b})
\end{equation}
\end{definition}

\begin{property}[Robustness Guarantee]
\label{theorem:robustness}
Let \(t\) be a prescribed safety threshold, and let \(\Omega(s_0)\) be a local perturbation domain for an input image \(s_0\). Suppose \(f_{OD}\) is an object detection model that, for every perturbed image in \(\Omega(s_0)\), outputs a predicted bounding box. 
Let us define the collection of 
all such bounding boxes as the set $\mathbf{b} = f_{OD}(\Omega(s_0))$.
Let us denote by \(b_{gt}\) the ground-truth bounding box for \(s_0\), and define
\[
\underline{IoU_{gt}}(\mathbf{b}) = \min_{b \in \mathbf{b}} IoU_{gt}(b),
\]
that is, the smallest intersection-over-union between \(b_{gt}\) and any box \(b\) in \(\mathbf{b}\). 
The model \(f_{OD}\) is said to be \textit{robust} to the perturbations in \(\Omega(s_0)\) if $\underline{IoU_{gt}}(\mathbf{b}) \geq t$.
\end{property}

This means that as long as the worst-case IoU remains above the threshold $t$, even under all perturbations, the detection is guaranteed to maintain sufficient overlap with the ground-truth box.
\comment{\begin{theorem}[Robustness guarantee]
\label{theorem:robustness}
If $t$, a safety threshold required by the system, the robustness of an object detection model to a perturbation domain $\Omega$ for an input image $s_0$ and its associated set of predicted bounding boxes $\mathbf{b}$ is guaranteed if $\underline{IoU_{gt}}(\mathbf{b}) \geq t$
\end{theorem} 
}

\comment{
    \min\limits_{\underline{f_{OD}}(\Omega)\; \leq b \leq\;  \overline{f_{OD}}(\Omega)} IoU_{gt}(b)\geq t
    \implies \min\limits_{ b \in  f_{OD}(\Omega)} IoU_{gt}(b)\geq t
}

\subsection{Overview of the proposed approach}
\begin{figure}[hbt]
\centering
\includegraphics[width=.8\linewidth]{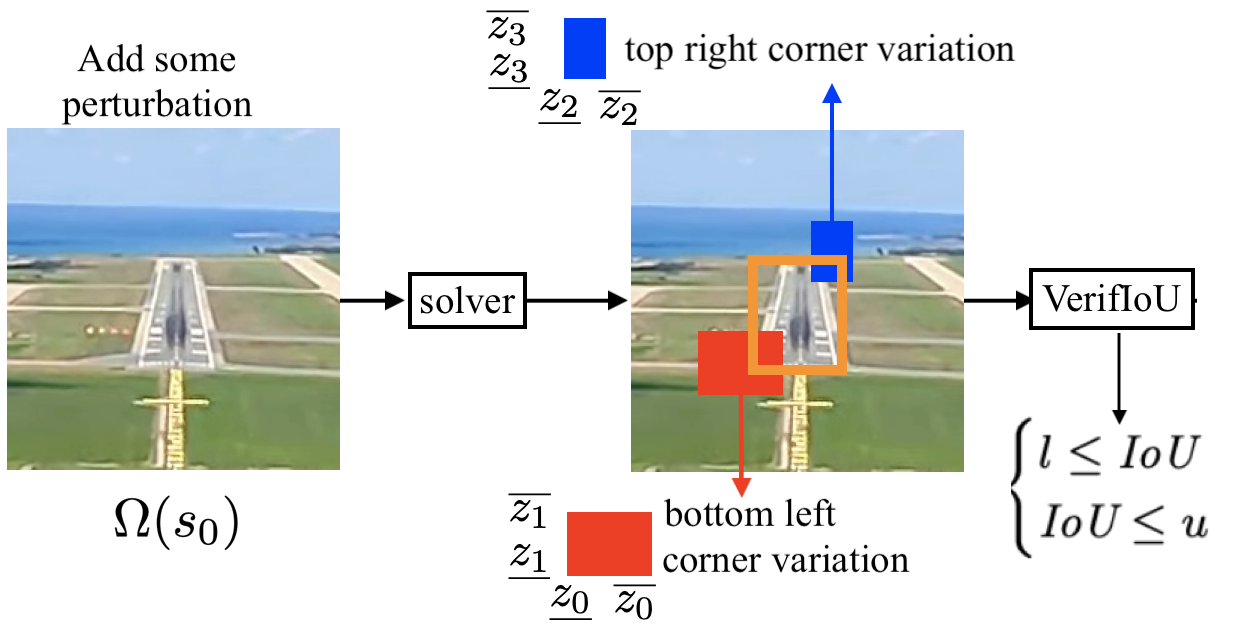}
\caption{Flow chart illustrating the two-step approach to derive bounds for the \iou for a specific test image $s_0$ and perturbation domain $\Omega$.}
\label{pipeline}
\end{figure}

An overview of the verification pipeline is shown in Figure~\ref{pipeline}.
Our approach is composed of two steps:
\begin{itemize}
\item Step 1: Compute $\mathbf{b} = f_{OD}(\Omega(s_0))$. Bounding boxes are defined by their bottom left ($z_0$,$z_1$) and upper right ($z_2$,$z_3$) coordinates i.e.,~an array of dimension $4$. Bounds of $\mathbf{b}$ therefore refer to bounds derived for these coordinates (see definition \ref{bbb}). We rely on verification tools based on abstract interpretation (e.g.~ERAN \cite{eranpaper}, Auto-LIRPA \cite{autolirpa} or decomon \cite{decomon}) to compute these bounds. 
\item Step 2: Compute bounds for the set of \iou corresponding to $\mathbf{b}$. We propose a novel approach and algorithm called \iouibp that relies on Interval Bound Propagation (IBP) \cite{mirman2018differentiable,gehr2018ai2}. As we will see, the key challenge lies in the non-linearity of the \iou metric.
\end{itemize}

The derived bounds of the \iou are then confronted to robustness requirements derived by system experts to assess whether the model is robust or not to perturbations e.g.~are all \iou above a robustness threshold of \iou $> 0.5$.

We emphasize that this 2-step verification is `incomplete' i.e, if the robustness is not guaranteed ($\underline{IoU_{gt}}(\mathbf{b}) < t$), it can either mean that the object detection model for a given input image is not robust to $\Omega$ or that the bounds derived by the verification solution might not be tight enough to provide guarantee.

\subsection{Detailed description of the approach}
For the Step 1 - Bounds for $\mathbf{b}$ using abstract interpretation,
we need to derive bounds for $\mathbf{b}$ i.e., $\underline{\mathbf{b}}$ and $\overline{\mathbf{b}}$.
We note that the detection model architecture must be compatible with the available abstract interpretation solvers; this, in particular, limits the choice of internal operators and how they are composed to obtain the candidate bounding box. More and more operators are luckily supported thanks to initiative such as \cite{autolirpa}. 

For the Step 2 - Bounds for \iou using IBP,
once we have bound estimates for $\mathbf{b}$, we then need to propagate them through the $IoU_{gt}$ function to derive $\underline{IoU_{gt}}(\mathbf{b})$ and $\overline{IoU_{gt}}(\mathbf{b})$. We rely on Interval Bound Propagation (IBP). IBP is a fast abstract interpretation approach that consists in propagation worst case constant bounds over a set of input intervals. 

Bounding $IoU_{gt}(\mathbf{b})$ is challenging as it is (i) a multi-dimensional input function (ii) neither convex, nor concave (iii) not piece-wise linear.
To tackle this problem, we explore two complementary approaches:
\begin{itemize}
    \item \textbf{{Vanilla\_IoU}} bounds the primitive operators and composes them using the rules of interval extension arithmetic (see section \ref{vanilla_iou}).
    \item \textbf{Optimal\_IoU } computes the highest and lowest $IoU_{gt}(\mathbf{b})$ values using some properties on the partial derivatives (see section \ref{optimal_iou}).
\end{itemize}

\subsubsection{Vanilla\_IoU - bounding the primitive operators}
\label{vanilla_iou}
We define the bounds derived for $IoU_{gt}(\mathbf{b})$ using the Vanilla\_IoU approach such that $IoU_{gt}(\mathbf{b}) \in [\underline{IoU}_v(\mathbf{b}),\overline{IoU}_v(\mathbf{b})]$.

\begin{figure*}[hbt]
\centering
\begin{tabular}{|>{\centering\arraybackslash}m{4cm}|>{\centering\arraybackslash}m{5cm}|>{\centering\arraybackslash}m{7cm}|}
\hline
\textbf{Object} & \textbf{Single Prediction} & \textbf{Bounds} \\
\hline 
Predicted bounding box & $b=[z_0, z_1, z_2, z_3]$& 
$\begin{aligned} 
\mathbf{b} &\subseteq& [\underline{\mathbf{b}}, \overline{\mathbf{b}}]\\
\underline{\mathbf{b}}&=&[\underline{z_0}, \underline{z_1}, \underline{z_2}, \underline{z_3}] \\
\overline{\mathbf{b}}&=&[\overline{z_0}, \overline{z_1}, \overline{z_2}, \overline{z_3}]
\end{aligned}$ \\
\hline 
Area of predicted bounding box & 
$a(b)= (z_2 - z_0) \times (z_3 - z_1)$
 &
$\begin{aligned}a(\mathbf{b}) &\subseteq& [\underline{a}(\mathbf{b}), \overline{a}(\mathbf{b})] \\
\underline{a}(\mathbf{b}) &=& (\underline{z_2}- \overline{z_0}) \times_{\geq 0} (\underline{z_3}- \overline{z_1}) \\
\overline{a}(\mathbf{b}) &=& (\overline{z_2}- \underline{z_0}) \times_{\geq 0} (\overline{z_3} - \underline{z_1})
\end{aligned}$
 \\
\hline
Intersection of predicted bounding box with ground truth {\small $\displaystyle
\begin{aligned}j=(gt,b)\end{aligned}$} {\small $\displaystyle
\begin{aligned} \underline{z_i}^{gt}=\overline{z_i}^{gt}\end{aligned}$}& $i_{gt}(b)=( \max z_0^j, \max
z_1^j, \min z_2^j, \min z_3^j )$
 & $i_{gt}(\mathbf{b}) \subseteq [\underline{{i}}(\mathbf{b}), \overline{{i}}(\mathbf{b})] $
 $\begin{aligned}
&\underline{{i}}(\mathbf{b})=( \max \underline{z_0}^j, \max \underline{z_1}^j, \min \underline{z_2}^j, \min \underline{z_3}^j ) \\
&\overline{{i}}(\mathbf{b})=( \max \overline{z_0}^j, \max \overline{z_1}^j, \min \overline{z_2}^j, \min \overline{z_3}^j ) \\
& where \max \underline{z_i}^j = \max (\underline{z_i},z_i^{gt}), i =0\dots3. 
\end{aligned}$
\\
\hline
$IoU_{gt}$ & $\iou_{gt}(b)=$
\[\frac{a(i_{gt}(b))}{a(b) + a(b_{gt}) - a(i_{gt}(b))}\]
 & $\iou_{gt}(\mathbf{b}) \subseteq [\underline{\iou}_v(\mathbf{b}), \overline{\iou}_v(\mathbf{b})]$
{\small $\displaystyle
\begin{aligned}
&\underline{\iou}_v (\mathbf{b}) = \dfrac{\underline{a}([\underline{{i}}(\mathbf{b}), \overline{{i}}(\mathbf{b})])}{\overline{a}(\mathbf{b}) + a( b_{gt}) - \underline{a}([\underline{{i}}(\mathbf{b}), \overline{{i}}(\mathbf{b})])} \\
&\overline{\iou}_v (\mathbf{b}) = \dfrac{\overline{a}([\underline{{i}}(\mathbf{b}), \overline{{i}}(\mathbf{b})])}{\underline{a}(\mathbf{b}) + a(b_{gt}) - \overline{a}([\underline{{i}}(\mathbf{b}), \overline{{i}}(\mathbf{b})])}
\end{aligned}$}
\\
\hline
\end{tabular}
\caption{\iou bound computation for a set of predicted bounding boxes}.
\label{table:comparison_iou_vanilla}
\end{figure*}

The \iou function is a combination of `primitive' functions: min, max, addition, subtraction, multiplication and division by a positive scalar. We extend traditional interval arithmetic (commonly used on point values) to closed intervals. We provide a reminder of the arithmetic interval for those operators in Appendix \ref{apprendix:intervalarithmetic} (excerpt from \cite{ancestralintervalarithmetic, ancestral2intervalarithmetic}). The expressions of $\underline{IoU}_v(\mathbf{b})$ and $\overline{IoU}_v(\mathbf{b})$
is defined in \fig~\ref{table:comparison_iou_vanilla}. Details on how these bound expressions were derived are provided in Appendix~\ref{apprendix:intervalarithmetic}.

\subsubsection{Optimal\_IoU extension - exact bounds}
\label{optimal_iou}
We define the bounds derived for $IoU_{gt}(\mathbf{b})$ using the Optimal\_IoU approach such that $IoU_{gt}(\mathbf{b}) \in [\underline{IoU}_{opt}(\mathbf{b}),\overline{IoU}_{opt}(\mathbf{b})]$.

\begin{figure*}[hbt]
   \begin{align}
\label{eq:derivative}
\begin{split}
     \frac{\partial IoU_{gt}(b)}{\partial z_{k=0,2}} = \frac{y_{max}-y_{min}}{d_{gt}(b)^2}\times\begin{cases}
     c_k (z_3-z_1)(x_{max}-x_{min}) \emph{ if } c_k z_k \leq c_k z_k^{gt}\\
     - c_k a(b_{gt}) + c_k(z_3-z_1)(x_{max}-z_2 + z_0 - x_{min}) 
    \end{cases}\,
\end{split}\\
\begin{split}
\label{eq:derivative2}
 \frac{\partial IoU_{gt}(b)}{\partial z_{k=1,3}} = \frac{x_{max}-x_{min}}{d_{gt}(b)^2}\times\begin{cases}
      c_k(z_2-z_0)(y_{max}-y_{min}) \emph{ if } c_k z_k \leq c_k z_k^{gt}\\
     - c_k a(b_{gt}) + c_k(z_2-z_0)(y_{max}-z_3 + z_1 - y_{min}) 
    \end{cases}\,
\end{split}
\end{align}
\caption{Equations for optimal\_IoU\label{fig:eq_optimal}}
\end{figure*}

We first derive the partial derivatives of $IoU_{gt}$ with respect to the predicted bounding boxes individual coordinates. These derivatives are shown in equations \ref{eq:derivative} and \ref{eq:derivative2} of figure \ref{fig:eq_optimal}. Details on how they were derived are provided in Appendix \ref{apprendix:partial_derivative}. For readability purposes, we introduce a few notations: 
\begin{itemize}
\item $x_{max}= \min(z_2, z_2^{gt})$ and $x_{min}= \max(z_0, z_0^{gt})$, 
\item $y_{max}=\min(z_3, z_3^{gt})$, $y_{min}= \max(z_1, z_1^{gt})$, 
\item $d_{gt}(b) = a(b_{gt}) + a(b) - a\big(i(b, b_{gt}) \big)$, 
\item $c_{k=2, 3}=-1$ and $c_{k=0, 1}=1$.
\end{itemize}

\iougt has the major advantage of having independent variations among its variables. This specificity allows us to optimize \iougt by coordinates and deduce the global optima of the interval extension function. The different variations of \iougt are depicted in \fig~\ref{tab:variations}. 

\begin{figure}[hbt]
\centering
\begin{tikzpicture}[scale=0.59]
   \tkzTabInit{$z$ / 1 , $\frac{\partial \iougt}{\partial z_0}$ / 1, $\frac{\partial \iougt}{\partial z_2}\;$ / 1}{$-\infty$, $z_0^{gt}$,$z_2^{gt}$, $\infty$}
   \tkzTabLine{, +, d, -, , -}
   \tkzTabLine{, +, , +, d,-}
\end{tikzpicture}
\begin{tikzpicture}[scale=0.59]
   \tkzTabInit{$z$ / 1 , $\frac{\partial \iougt}{\partial z_1}$ / 1, $\frac{\partial \iougt}{\partial z_3}\;$ / 1}{$-\infty$, $z_1^{gt}$,$z_3^{gt}$, $\infty$}
   \tkzTabLine{, +, d, -, , -}
   \tkzTabLine{, +, , +, d,-}
\end{tikzpicture}
\caption{Variation of the partial derivatives of \iougt
\label{tab:variations}}
\end{figure}

The $+$/$-$ signs indicate that the derivative is increasing/decreasing over the interval, independently of the other coordinates. \iougt is increasing when the input variables get closer to the ground truth coordinates $b_{gt}=[z^{gt}_i]_{i=0}^3$. 

Computing the coordinates of the most optimal box $b_u^*$ (i.e.,~the predicted bounding box that will provide the highest value of \iougt) is immediate:
\begin{align}\label{equ:optimal_upper_iou}
\begin{split}
\overline{IoU}_{opt}(\mathbf{b}) &= \max\limits_{{b}\in [\underline{\mathbf{b}}, \overline{\mathbf{b}}]} IoU_{gt}({b}) \\
& = IoU(b_u^*=[z^*_i]_{i=0}^3,  b_{gt}=[z^{gt}_i]_{i=0}^3 )\\
&\hspace{1cm}
\text{ with}\;  z^*_i = \begin{cases}
      z_i^{gt} \; &\text{ if } z_i^{gt} \in [\underline{\mathbf{b}}_i, \overline{\mathbf{b}}_i]\\
      \underline{\mathbf{b}}_i &\text{ if } z_i^{gt} \leq \underline{\mathbf{b}}_i\\
      \overline{\mathbf{b}}_i &\text{ else}\\
    \end{cases}\,
\end{split}
\end{align}
i.e.,~select for $z_i^*$ the coordinate of the ground truth $z_i^{gt}$ if it belongs to the interval $z_i = [\underline{z_i}, \overline{z_i}]$; otherwise, choose the lower bound if $z_i^{gt}$ is on the left of $z_i$, or the upper bound if $z_i^{gt}$ is on the right. \fig~\ref{ioualignment} shows a visual representation of the choice of $z_i^*$.

\newsavebox{\figureA}
\savebox{\figureA}{%
\begin{tikzpicture}
    \draw[->] (-0.7,0) -- (6,0);
    \draw[thick] (4,-0.1) -- (4,0.1) node[above] {$\overline{z_i}$};
    \draw[thick] (1,-0.1) -- (1,0.1) node[above] {$\underline{z_i}$};
    \draw[<->,red,line width=0.4mm] (4,0) -- (1,0);
    \draw[fill=black] (2.5,0) circle (0.05) node[below] {$z_i^{gt}$};
\end{tikzpicture}
}

\newsavebox{\figureB}
\savebox{\figureB}{%
\begin{tikzpicture}
    \draw[->] (-0.7,0) -- (6,0);
    \draw[thick] (4,-0.1) -- (4,0.1) node[above] {$\overline{z_i}$};
    \draw[thick] (1,-0.1) -- (1,0.1) node[above] {$\underline{z_i}$};
    \draw[<->,red,line width=0.4mm] (4,0) -- (1,0);
    \draw[fill=black] (0,0) circle (0.05) node[below] {$z_i^{gt}$};
\end{tikzpicture}
}

\newsavebox{\figureC}
\savebox{\figureC}{%
\begin{tikzpicture}
    \draw[->] (-0.7,0) -- (6,0);
    \draw[thick] (4,-0.1) -- (4,0.1) node[above] {$\overline{z_i}$};
    \draw[thick] (1,-0.1) -- (1,0.1) node[above] {$\underline{z_i}$};
    \draw[<->,red,line width=0.4mm] (4,0) -- (1,0);
    \draw[fill=black] (5,0) circle (0.05) node[below] {$z_i^{gt}$};
\end{tikzpicture}
}

\begin{figure}[hbt]
\centering
\scalebox{0.95}{
\begin{tabular}{|M{7.2cm}|M{0.8cm}|}
\hline
Representation of distance from $z_i$=[$\underline{z_i}$, $\overline{z_i}$] to $z_i^{gt}$ & \textbf{ $z_i^*$} \\
\hline
\usebox{\figureA} & $z_i^{gt}$ \\
\hline
\usebox{\figureB} & $\underline{z_i}$ \\
\hline
\usebox{\figureC} & $\overline{z_i}$ \\
\hline
\end{tabular}}
\comment{\begin{minipage}{0.29\linewidth}
 To compute the optimal maximum \iougt, select the ground truth coordinate if it belongs to the interval; otherwise, choose the lower bound if the coordinate is on the left, or the upper bound if it is on the right.
\end{minipage}}
\caption{Selection of coordinates providing the highest \iougt}
\label{ioualignment}
\end{figure}

The interval extension of $\mathcal{B}$ is naturally defined as the joint product of the interval extension for each coordinate of a box. However, this definition can lead to ill defined boxes, as the coordinates of the upper right corner may be lower than those of the bottom left corner. Those corner cases happen whenever $\underline{z}_2\leq \overline{z}_0$ or $\underline{z}_3\leq \overline{z}_1$. Corner cases create an infinite number of collapsed bounding boxes $\{[z_0, z_1, z_0, z_3]\}\cup \{[z_0, z_1, z_2, z_1]\}$ whose $IoU_{gt}$ saturates to 0. 

Let's look into the lowest possible value for \iougt i.e.,~the most relevant bound of this analysis as it will define if the model we are testing is robust or not to perturbation for a certain test image. Since \iougt is decreasing when an input variable is getting away from the ground truth coordinates, the predicted bounding box that will generate the lowest \iougt is one of the vertices of the input domain of \iougt: 
\begin{align}
\label{equ:optimal_lower_iou}
\begin{split}
\underline{IoU}_{opt} (\mathbf{b}) &= \min\limits_{{b}\in [\underline{\mathbf{b}}, \overline{\mathbf{b}}]} IoU_{gt}(b) \\
&= \begin{cases}
      0 \; &\text{ if } \underline{z}_2\leq \overline{z}_0 \text{ or } \underline{z}_3\leq \overline{z}_1 \\
      \min\limits_{{b} \in \{\underline{\mathbf{b}}, \overline{\mathbf{b}}\}} IoU_{gt}({b}) &\text{ otherwise } \\
    \end{cases}\,
\end{split}
\end{align}


Let's now illustrate and benchmark the different step 1 and step 2 bound derivation approaches on a couple of use cases and perturbations.

\section{Experiments} \label{sec:experiments}

\noindent {\bf Setting:} The verification approach is tested on two object detection use cases and a number of perturbations. We focus on CNN-based object detection models and benchmark a number of techniques for step 1 along with the two implemented solutions for step 2. The code (in python) is made available on github\footnote{\url{https://github.com/NoCohen66/Verification4ObjectDetection}}.
All experiments are in part parallelized over a pool of $20$ workers, on a Linux machine with Intel\textsuperscript{\textregistered}  Xeon\textsuperscript{\textregistered} processor E5-2660 v3 @ 2.60GHz of 20 cores and 64 GB RAM. \\

\noindent {\bf Object detection use cases:} We explore two datasets, namely:
\begin{itemize}
    \item \textbf{\dlocmaj}: the localization of handwritten digit randomly placed on black background images. The digits originate from the grey-scale MNIST dataset \cite{MNIST}. MNIST images of size $28\times28$ are randomly placed on black images of size of $90\times90$, thus the ground truth box has a fixed size on all images and its coordinates correspond to the position of the MNIST image. We train a CNN digit detection model, whose description is provided in \fig~\ref{tab:archi}. It outputs $4$ values that predicts the four coordinates. The verification is conducted on $40$ images. The generated dataset is uniformly composed of images representing digits from $0$ to $9$.
    
    \item \textbf{LARD}: the LARD dataset \cite{LARD}. LARD comprises high-quality aerial images of runway during approach and landing phases. We train a CNN runway detection model, whose description is provided in \fig~\ref{tab:archi}. We select 40 synthetic images from the Reykjavík domestic airport taken into clear weather conditions within a distance range of $0.33$ to $1.08$ nautical miles (NM) from the runway that are resized to a size of $256 \times 256$ pixels, with runway projected sizes ranging between $70$ and $706$ pixels. This second use case is more challenging for robustness verification due to varying ground truth box sizes.
\end{itemize}

We normalize pixel intensity values to a scale ranging from 0 to 1.

\begin{figure}[hbt]
    \centering
        \begin{tabular}{c|c}
			\dlocmaj \text{  }  \textsc{CNN} & \textsc{LARD CNN}\\
   \hline
        \textsc{Conv} 16 3$\times$3/1/1 - \textsc{ReLU} &  \textsc{Conv} 32 3$\times$3/2/1 - \textsc{ReLU} \\
        \textsc{Pool} 2$\times$2/2 - \textsc{ReLU}& \textsc{Conv} 64 3$\times$3/2/1 - \textsc{ReLU} \\
        \textsc{Conv} 16 3$\times$3/1/1 - \textsc{ReLU}& \textsc{Conv} 128 3$\times$3/2/1 - \textsc{ReLU}\\
        \textsc{Pool} 2$\times$2/2 - \textsc{ReLU}& \textsc{Flatten}\\
        \textsc{Flatten} & \textsc{Linear} 128 - \textsc{ReLU}\\
        \textsc{Linear} 256 - \textsc{ReLU}& \textsc{Linear} 128 - \textsc{ReLU}\\
        \textsc{Linear} 4 & \textsc{Linear} 4 
    \end{tabular}
    \caption{Overview of network architectures. \textsc{Conv} c h$\times$w/s/p corresponds to a 2D-convolution 
    with c output channels, h$\times$w kernel size, 
    stride s in both dimensions, padding p. 
    Pooling layers are specified analogously \label{tab:archi}}
\end{figure}

\begin{figure*}[hbt]
\centering
\begin{tabular}{c|c|c|ccc|ccc}
Perturbation & Factor &  $\Omega(s_0)$ & \multicolumn{3}{c|}{\dlocmaj \text{  }} & \multicolumn{3}{c}{\textsc{LARD}} \\
(1) & (2) & (3) & \multicolumn{3}{c|}{(4)} & \multicolumn{3}{c}{(5)} \\
\hline
&&& min& max&step&min&max&step\\
White noise & $\epsilon$ &$\{s \in \mathbb{R}^n \mid||x-s_0||_\infty \leq \epsilon \}$& 0 & 0.002&11&0&0.002&11 \\
Brightness & $\alpha_b$ &$\{s \in \mathbb{R}^n \mid s= s_0+\alpha_b \}$& 0 & 0.002&11&0&0.02&11 \\
Contrast & $\alpha_c$ &$\{s \in \mathbb{R}^n \mid s= s_0\times\alpha_c \}$& 0 & 0.2&11&0&0.1&11 \\
\end{tabular}
\caption{Tested perturbation intensities \label{fig:param-perturbation}}
\end{figure*}

\noindent {\bf Perturbations:} We explore three types of perturbations: white noise, brightness and contrast. White noise naturally occurs in video recording due to e.g.~sensor sensitivity. Contrast and brightness are also naturally impacting images e.g.,~when captured under challenging weather conditions or time of day. A noise perturbation domain consists of all images potentially obtained by applying an additive value to each pixel independently. The value of noise is usually limited to a certain threshold ($\pm \epsilon$). Brightness/contrast perturbation domains consist of all images obtained by applying a uniform additive/multiplicative coefficient $\alpha_b$/$\alpha_c$, respectively. 

\fig~\ref{fig:param-perturbation} summarises the perturbation domain definitions (column 3) and tested perturbation intensities (columns 4 \& 5). For white noise, we thus consider images whose pixels are  affected individually by $11$ incremental perturbation domains with $\epsilon = 0$, $|\epsilon| \leq 0.0002$, ..., $|\epsilon| \leq 0.002$. For contrast (and LARD), we consider $11$ incremental ranges of $\alpha_c$ around 0 with $\alpha_c = 0$, $\alpha_c \in [-0.01,0.01]$, ..., $\alpha_c \in [-0.1,0.1]$. 
\\



\noindent {\bf Benchmarked techniques:} For step 1, the bounds $\mathbf{\interval{b}}$ are obtained using the Auto-LiRPA verification tool \cite{autolirpa}. We consider three verification methods: IBP \cite{IBP}, CROWN-IBP \cite{zhang2019towards}, and CROWN \cite{CROWN}. For step 2, we benchmark the two approaches Vanilla\_IoU and Optimal\_IoU. \\

\noindent {\bf Robustness metric:} To compare the efficiency of the different (combination of) verification techniques, we introduce the notion of \textit{Verified Box Accuracy} (VBA) that corresponds to the fraction of test images fulfilling the robustness guarantee property from theorem \ref{theorem:robustness} with a threshold $t = 0.5$.

\section{Results} \label{sec:results}

\begin{figure}[!hbt] 
\centering
\begin{tabular}{lcc}
 \includegraphics[width=0.5\linewidth]{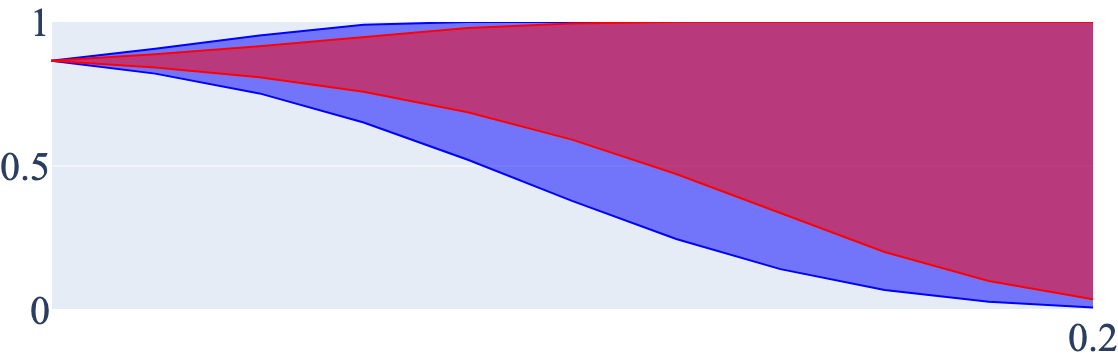}
\end{tabular} 
\caption{Average bounds for \iou (y-axis) for increasing contrast perturbation (x-axis) on the two test dataset. Bounds derived with Optimal\_IoU and Vanilla\_IoU are shown in red and blue respectively, on DIGIT$\_$LOC dataset.}
\label{fig:averageioumnistlard}
\end{figure}

Figure~\ref{fig:averageioumnistlard} shows the (average) \iou bounds derived on the test images for the two use cases, the three investigated perturbations and the two tested solutions for step 2. Results are shown for experiments using CROWN as step 1. When no perturbation is applied (i.e.,~perturbation intensity $= 0$), the \iou is represented by a single value. $IoU > 0.5$ as we are only considering test images with a good detection. As the perturbation intensity increases, the \iou is represented by its corresponding bounded interval with widening bounds. Some of the results are provided in \fig~\ref{tab:parameters} for illustration.

\begin{figure}[!hbt]
\centering
\begin{tabular}{|l|c||c|c|c||c|c|c|}
\hline
\multicolumn{2}{|c||}{} & \multicolumn{3}{c||}{\dlocmaj \text{ }} & \multicolumn{3}{c|}{LARD} \\
\hline
\hline
\multicolumn{2}{|c||}{\textbf{White noise} $\epsilon =$ } &  $0.0002$ & $ 0.0004$ & $ 0.0006$ & $0.0004$ & $0.0006$ & $ 0.0008$ \\
\hline
\multirow{2}{*}{IoU$_v$} 
& C-IBP & 76.9 & 0.0 & 0.0 & 0.0 & 0.0 & 0.0 \\
& C &    97.4 & 0.0 & 0.0 & 25.0 & 2.8 & 0.0 \\
\hline
\multirow{2}{*}{IoU$_{opt}$}
& C-IBP & 100.0 & 35.9 & 2.6 & 8.3 & 0.0 & 0.0 \\
& C &  100.0 & 66.6 & 7.70 & 97.2 & 75.0 & 27.8 \\
\hline
\hline
\multicolumn{2}{|c||}{\textbf{Brightness} $\alpha_b = $} & $0.0002$ & $0.0004$ & $0.001$ & $ 0.004$ & $ 0.006$ & $0.008$ \\
\hline
\multirow{2}{*}{IoU$_v$}
& C-IBP   & 87.2 & 0.0 & 0.0 & 0.0 & 0.0 & 0.0 \\
& C &    100.0 & 100.0 & 17.9 & 77.8 & 38.9 & 11.1 \\
\hline
\multirow{2}{*}{IoU$_{opt}$} 
& C-IBP   & 100.0 & 35.0 & 0.0 & 0.0 & 0.0 & 0.0 \\
& C &   100.0 & 100.0 & 92.3 & 94.4 & 86.1 & 66.7 \\ 
\hline
\hline
\multicolumn{2}{|c||}{\textbf{Contrast} $\alpha_c = $} &   $0.0002$ & $ 0.001$ & $ 0.0014$ & $0.01$ & $ 0.02$ & $ 0.03$ \\
\hline
\multirow{2}{*}{IoU$_v$} 
& C-IBP   & 31.4 & 0.0 & 0.0 & 0.0 & 0.0 & 0.0 \\
& C &    100.0 & 8.6 & 0.0 & 69.1 & 32.0 & 13.4 \\
\hline
\multirow{2}{*}{IoU$_{opt}$}
& C-IBP &  82.9 & 0.0 & 0.0 & 0.0 & 0.0 & 0.0 \\
& C &  100.0 & 88.6 & 8.6 & 82.5 & 58.8 & 38.1 \\
\bottomrule
\end{tabular}
\caption{Examples of VBA (in \%) obtained for some of the tested perturbations for different verification approach combinations and the two use cases.}
\label{tab:parameters}
\end{figure}

\noindent We observe:

\begin{itemize}

\item {\bf the importance of the choice of solver for step 1:} For all dataset, at fixed solution for step 2 and fixed perturbation intensity, the VBA is systematically smaller for experiences using CROWN-IBP versus CROWN. We find for example a VBA of $76.9\%$ vs. $97.4\%$ for CROWN-IBP versus CROWN, for the DIGIT\_LOC dataset, a noise of $\epsilon=2\times 10^{-4}$ and a Vanilla\_IoU for step 2. We thus observe the importance of the tightness of CROWN vs CROWN-IBP. We also note that using a pure IBP approach for step 1 always results in a VBA of $0$ i.e.,~fails to converge into any robustness guarantees.
 
\item {\bf the higher efficiency of Optimal\_IoU:} vs. Vanilla\_IoU approach in providing guarantees. We observe in Fig.~\ref{fig:averageioumnistlard} that the \textit{envelope} created by the bounds derived using Optimal\_IoU (red) is tighter that the one derived for Vanilla\_IoU (blue), This figure shows the overapproximation made by Vanilla\_IoU. \comment{This difference translates to a \emph{false positive rate} ranging from $2.6\%$$ to $80\%$$ for the standard threshold (t=0.5).} In Fig.~\ref{tab:parameters}, we see that the VBA metric is systematically higher for Optimal\_IoU. We find for example a VBA of $25.0\%$ vs. $97.2\%$ for Vanilla\_IoU vs. Optimal\_IoU, for the LARD test dataset, a white noise of $\epsilon=2\times 10^{-4}$ and CROWN for step 1. These results are showcasing that the Optimal\_IoU approach is able to derive tighter bounds for the \iou and to provide safety guarantees for a larger number of test images.
\end{itemize}

\begin{figure}[hbt] 
\centering
\includegraphics[width=0.5\linewidth]{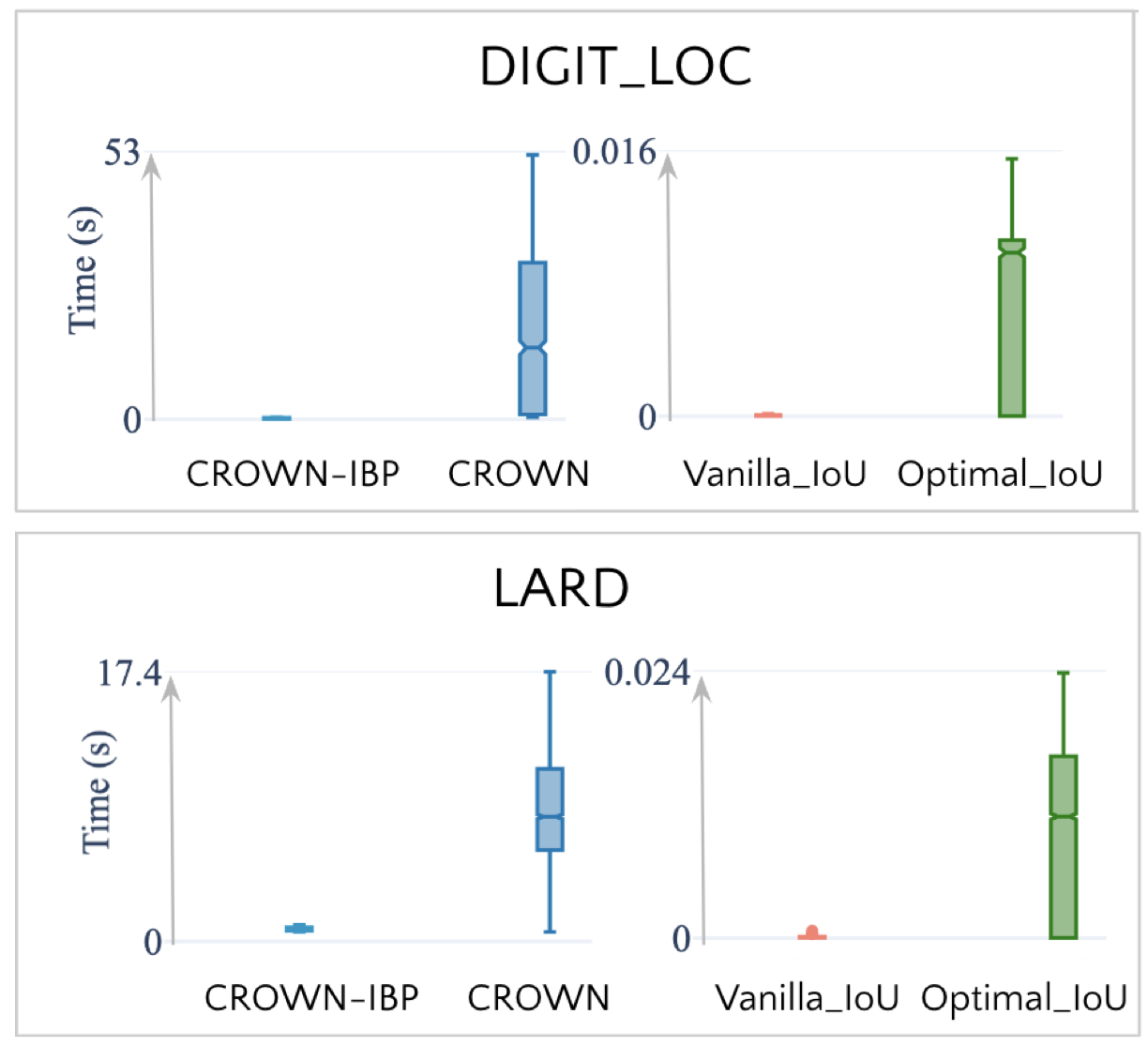}
\caption{Computation times for step 1 and 2.}
\label{fig:timeLARDandMNIST}
\end{figure}

Figure~\ref{fig:timeLARDandMNIST} provides some insights into the computation time required for step 1 and 2. Unsurprisingly, we observe that Optimal\_IoU is a more computationally-heavy approach that Vanilla\_IoU. The computation time required for the step 2 calculations is however comparatively small compared to step 1.\\

In subsequent work, Raviv et al. \cite{raviv2024formal} introduced an approach that applies abstract interpretation to the primitive operators of the Intersection over Union (IoU) metric. A comparison with our method—which is restricted to verifying whether a property is satisfied—demonstrates that our approach substantially outperforms theirs. Specifically, in the reported whitenoise setting, their method achieves a VBA of 27\%, whereas our approach attains a VBA of 35\% on the \textbf{\dlocmaj} use case.

\noindent {\bf A special focus on the LARD use case: }The experiments show a dependency of the model robustness to perturbation with the size of the runway (a.k.a distance of the plane to the runway) with images with smaller runways showing more vulnerabilities to perturbations. 

\begin{figure}[hbt]
\centering
\begin{subfigure}{0.48\linewidth}
    \centering
    \includegraphics[width=\linewidth]{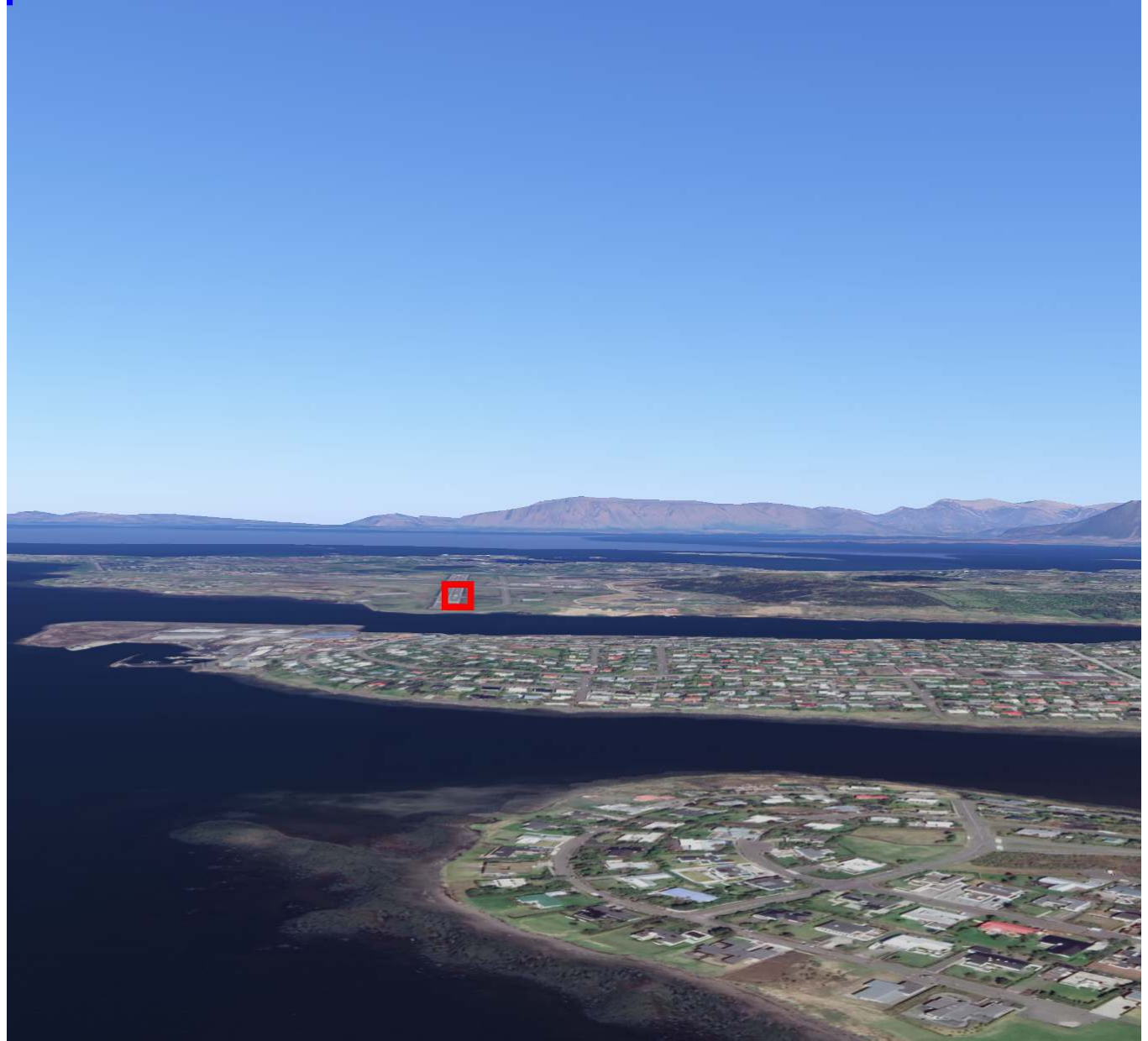}
    \caption{3.15km/1.70NM}
    \label{fig:3.15km}
\end{subfigure}
\hfill
\begin{subfigure}{0.48\linewidth}
    \centering
    \includegraphics[width=\linewidth]{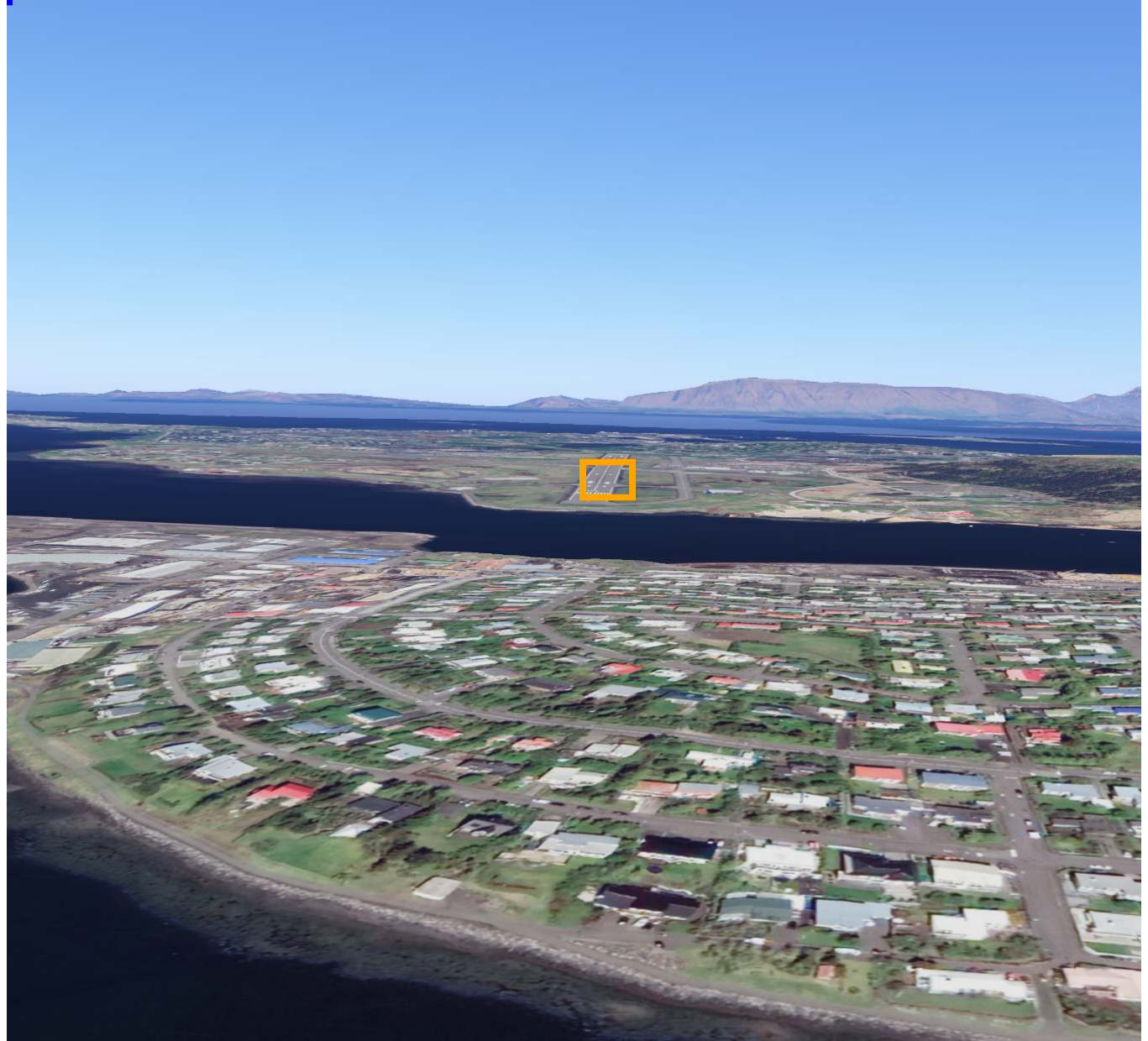}
    \caption{2.0km/1.08NM}
    \label{fig:2.0km}
\end{subfigure}

\vspace{2mm}

\begin{subfigure}{0.48\linewidth}
    \centering
    \includegraphics[width=\linewidth]{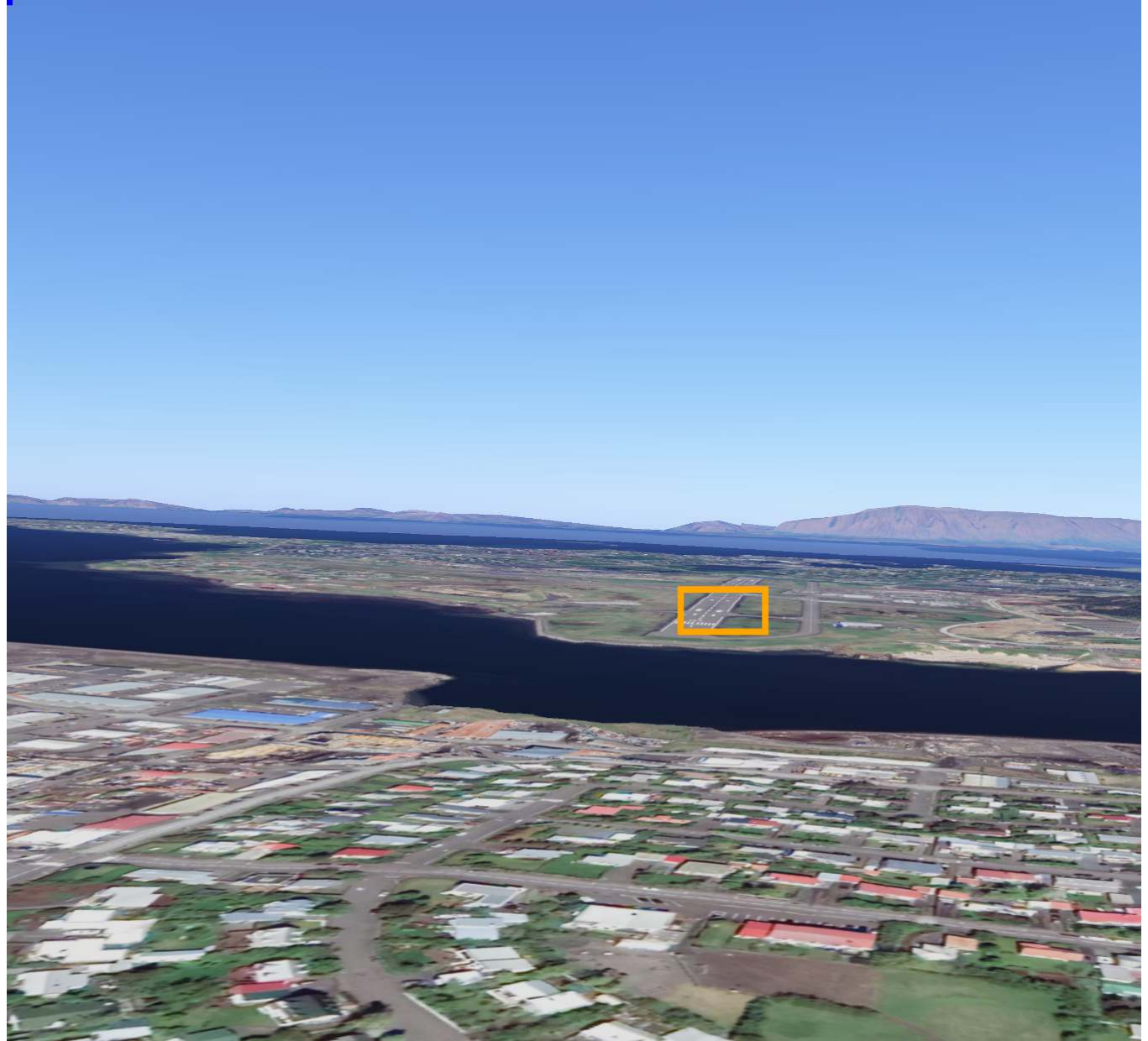}
    \caption{1.63km/0.88NM}
    \label{fig:1.63km}
\end{subfigure}
\hfill
\begin{subfigure}{0.48\linewidth}
    \centering
    \includegraphics[width=\linewidth]{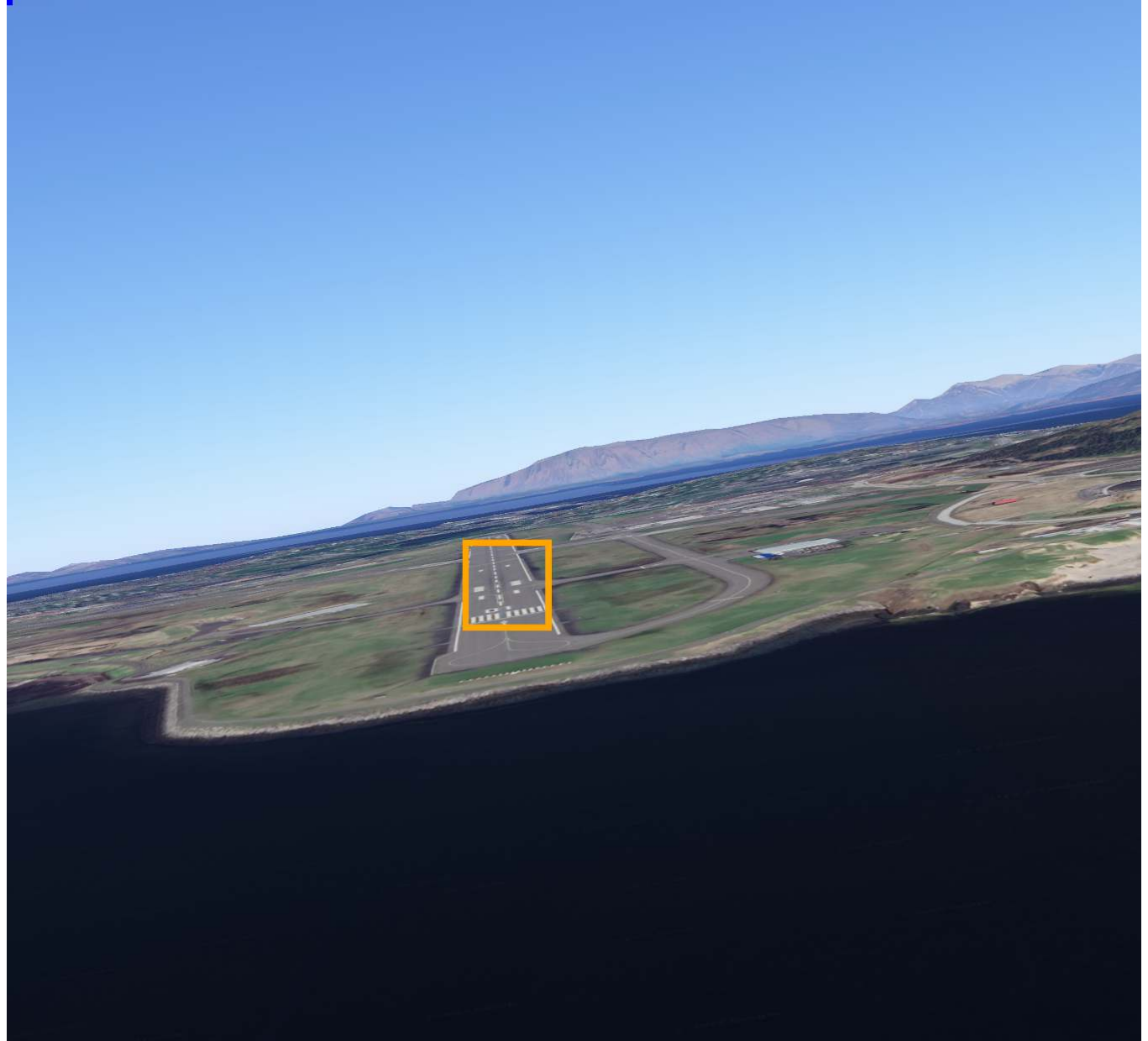}
    \caption{0.78km/0.42NM}
    \label{fig:0.78km}
\end{subfigure}

\caption{Impact of a brightness perturbation $\alpha_b = 0.002$ on a trajectory computed with the CROWN method: orange boxes indicate slight impact, while the red box indicates a strong one (with a minimal value of $\underline{IoU}_{opt}$ of 0.34).}
\label{fig:impact_of_brightness_perturbation}
\end{figure}

Figure~\ref{fig:impact_of_brightness_perturbation} showcases four images extracted from one landing approach for which we evaluate the robustness to a brightness perturbation domain of $\alpha_b = 0.002$.


This dependency is not too surprising as small impact on objects with small amount of pixels will have larger consequence on the \iou derivation than small impact on large objects. It demonstrates however the added challenge that AI practitioners face while training models with a range of object size and additional care they will have to dedicate to make their model robust across the whole size range.

\noindent {\bf Robustness training: }The ultimate objective for this work would be to seamlessly integrate this type of robustness evaluation while training a model, balancing performance on the \iou metric and robustness to perturbation on the training samples. Our effort to reduce the computational cost of the solution implementation goes in the right direction. \comment{Future work will consider this type of certified training for object detection. }

\section{Conclusion} \label{sec:conclusions}

We present a novel approach to the formal verification of object detection models. Our main contribution lies in the formalisation of non-linear, single box, robustness property, which allow the evaluation of the robustness of a detection model to local perturbations. 

The key idea is to bound the most extreme values of the \iou, the commonly-adopted performance metric for detection models. We remind that the \iou is multi-dimensional, non convex/concave and without an inherent property of partial monotonicity. To enable this, we first derive the impact of the perturbations on the bounding boxes outputed by the models using classically-used abstract interpretation techniques. We then propagate intervals through the \iou function, following two approaches: (1) bounding the primitive operators (Vanilla\_IoU), (2) applying interval extension on the \iou function (Optimal\_IoU). Optimal\_IoU offers a precise and fast formulation that is agnostic to both the network architecture and the type of local perturbation, as long as the ground-truth box remains fixed. Bringing it fully into real-world use now mainly requires extending the benchmark to include a wider range of plausible perturbations.

\comment{As future work, we will continue addressing formal verification of object detection by considering more operators such as Non-maximum Suppression (NMS) (this will enable us to extend IBP IoU to multiple bounding boxes and multi-class object detection) and more classical object detection models such as YoLo.
These formal methods aim at contributing to the certification in general.
Thus, we also would like to define the system expected properties. In particular, looking at the approach of \fig~\ref{fig:impact_of_brightness_perturbation},
an open question is: at which distance do we expect the object detection to be robust? and with which threshold?
As we already mentioned in the experiments section, we would also like to study the open question of certified training for object detection.}

\section*{Acknowledgment}
Our work has benefitted from the AI Interdisciplinary Institute ANITI. ANITI is funded by
the France 2030 program under the Grant agreement n°ANR-23-IACL-0002.
\bibliographystyle{abbrv}
\bibliography{bibliography}
\section{Appendices: Interval arithmetic operations}
\label{apprendix:intervalarithmetic}

This appendix presents the interval arithmetic formulas underlying the bound computations of the Vanilla\_\iou method. 

\begin{figure}[hbt]
    \centering
    \begin{tabular}{|c|c|c|}
    \hline
    \textbf{Notation} & \textbf{Formula} \\
    \hline
    +& $[\underline{a}, \overline{a}] + [\underline{b}, \overline{b}] = [\underline{a}+\underline{b}, \overline{a}+\overline{b}]$\\
    \hline
    -& $[\underline{a}, \overline{a}] - [\underline{b}, \overline{b}] = [\underline{a}-\overline{b}, \overline{a}-\underline{b}]$\\
    \hline
    $\times_{\geq 0}$ &  $[\underline{a}, \overline{a}] \cdot [\underline{b}, \overline{b}] = [\underline{a} \cdot \underline{b}, \overline{a} \cdot \overline{b}]$\\
    \hline
    $/$& $\dfrac{1}{[\underline{a}, \overline{a}]}= [\dfrac{1}{\overline{a}}, \dfrac{1}{\underline{a}}]$\\
    \hline
    $\min$ & $\max([\underline{a},\overline{a}], [\underline{b},\overline{b}]) = [\max(\underline{a}, \underline{b}), \max(\overline{a}, \overline{b})]$\\
    \hline
    $ \max$ & $\min([\underline{a},\overline{a}], [\underline{b},\overline{b}]) = [\min(\underline{a}, \underline{b}), \min(\overline{a}, \overline{b})]$\\
    \hline
    \end{tabular}
    \caption{Interval arithmetic operations: addition, subtraction, positive multiplication, positive division, minimum, maximum. }
    \label{tab:interval_summary}
\end{figure}

For example, for the area $a(\mathbf{b})$ of the predicted box:
$a(\mathbf{b}) \in ([\underline{z_2},\overline{z_2}] - [\underline{z_0},\overline{z_0}]) \times ([\underline{z_3},\overline{z_3}] - [\underline{z_1},\overline{z_1}])$, by first subtracting the corresponding intervals and then applying positive multiplication, we get:

$a(\mathbf{b}) \in ([\underline{z_2} - \overline{z_0} , \overline{z_2} - \underline{z_0}]) \times ([\underline{z_3} - \overline{z_1} , \overline{z_3} - \underline{z_1}])$,

$a(\mathbf{b}) \in [(\underline{z_2} - \overline{z_0}) \times_{\geq 0} (\underline{z_3} - \overline{z_1}) , (\overline{z_2} - \underline{z_0}) \times_{\geq 0} (\overline{z_3} - \underline{z_1})]$. This gives $\underline{a}(\mathbf{b})$ and $\overline{a}(\mathbf{b})$ as: $\underline{a}(\mathbf{b}) = (\underline{z_2}- \overline{z_0}) \times_{\geq 0} (\underline{z_3}- \overline{z_1})$ and $\overline{a}(\mathbf{b}) = (\overline{z_2}- \underline{z_0}) \times_{\geq 0} (\overline{z_3} - \underline{z_1})$.

\section{Appendices: Partial derivatives}
\label{apprendix:partial_derivative}

\begin{hypothesis}\label{hyp:not0}We only consider cases where the ground truth bounding box and the predicted bounding boxes for a perturbation domain $\Omega$ overlap.\end{hypothesis}

The partial derivative of $IoU_{gt}$ with respect to $z_i$ is derived using the quotient rule and the derivative of the maximum function such that, $\forall z_{i} \in \mathbb{R}$:
\begin{equation}
    \dfrac{\partial IoU_{gt}(b)}{\partial z_{i}} = \dfrac{ d_{gt}(b) \cdot \displaystyle 
        \dfrac{\partial a(i_{gt}(b))}{ \partial z_{i}} 
      - a(i_{gt}(b)) \cdot  \dfrac{\partial d_{gt}(b)}{\partial z_{i}}   }{d_{gt}(b)^2}
\label{f/g}
\end{equation}

where $d_{gt}(b) = a(b_{gt}) + a(b) - a\big(i_{gt}(b)\big)$ (the \iou denominator).

For instance, consider the partial derivative with respect to $z_0$, it can be written as: 

$a\big(i_{gt}(b)) = {E_0} \cdot (\min(z_2, z_2^{gt}) - \max(z_0, z_0^{gt}))$ 

$d_{gt}(b) = E_1 - z_0 \cdot E_2  -  a\big(i_{gt}(b))$ 

where $E_{0-2}$ are the positive canonical forms enumerated in Fig.~\ref{tab:constants} (top). $E_{0-5}$ are independent of $z_0$ and positive (see Fig.~\ref{tab:constants}, bottom).

Two cases arise:
\begin{itemize}
    \item Case A: if $z_0 < z_0^{gt}$, $a\big(i_{gt}(b)\big)$ = $E_0\cdot E_3$ and $d_{gt}(b) =  E_1 - z_0 \cdot E_2  -  E_0\cdot E_3$
    \item Case B: if $z_0 > z_0^{gt}$,  $a\big(i_{gt}(b)\big) = E_0 \cdot  (\min(z_2, z_2^{gt})-z_0)$ and $d_{gt}(b) = (E_0-E_2)z_0 + (E_1 - E_0 E_3)$.
\end{itemize}


\begin{figure}[hbt]
\centering 
\setlength{\extrarowheight}{-15pt} 
\renewcommand{\arraystretch}{0.4}

\begin{tabular}{ |p{1cm}||p{4cm}|p{2.4cm}|  }
\hline
\textbf{Name} & \textbf{Value} & \textbf{$E_i$ positivity justification}\\
\hline
$E_0$     & $\min(z_3, z_3^{gt}) - \max(z_1, z_1^{gt})$ & Eq. \eqref{rule_coord1}\\
$E_{1}$   &   $z_2 \cdot(z_3 - z_1) + a(b_{gt})$ & Eq. \eqref{rule_coord1} \\
$E_2$     &  $z_3 - z_1$ & Eq. \eqref{rule_coord1} \\
$E_3$     & $\min(z_2, z_2^{gt}) - z_0^{gt}$ & Eq. \eqref{rule_coord1} and \eqref{rule_coord2bis} \\
$E_{5}$   &  $z_2-\min(z_2, z_2^{gt})$ & Eq. \eqref{rule_coord4} \\
\hline
\end{tabular}
\vspace{0.1cm}
 
\begin{tabular}{|m{4cm} | m{4cm}|}
\hline
\textbf{Equation} & \textbf{Justification} \\
\hline
\begin{equation} a - \min(b,a) \geq 0 \label{rule_coord4} \end{equation} & Non negative difference \\ 
\hline 
By definition: 
\begin{equation}z_0 \leq z_2  \text{ and } z_1 \leq z_3 \label{rule_coord1} \end{equation} &  
\begin{tikzpicture}[thick,scale=0.75, every node/.style={scale=1}] 
\fill[red!40!white] (0,0) rectangle (2,1); 
\node at (0.5,0) [below left] {$(z_0, z_1)$}; 
\node at (1.5,1) [above right] {$(z_2, z_3)$}; 
\node at (0,0) {x};
\node at (2,1) {x};
\end{tikzpicture} \\ 
\hline 
\text{Under Hypothesis~\ref{hyp:not0}:}
\begin{equation} z_0 \leq z_0^{gt}  \implies z_2 \geq z_0^{gt} \label{rule_coord2} \end{equation}
\begin{equation} z_2 \leq z_2^{gt}  \implies z_2 \geq z_0^{gt} \label{rule_coord2bis} \end{equation}&
\begin{tikzpicture}[thick,scale=1, every node/.style={scale=1}] 

        \fill[red!40!white] (2,2.5) rectangle (0.5,1.5); 
        \fill[green!40!white] (3,2) rectangle (1,0.5); 
        \fill[orange!40!white] (2,2) rectangle (1,1.5); 
        \node at (2,2.5) {x}; 
        \node at (2,2.8) {$z_2$}; 
        \node at (0.5,1.5) {x}; 
        \node at (0.5,1.2) {$z_0$}; 
        \node at (3,2.35) {$z_2^{gt}$};
        \node at (3,2) {x};
        \node at (1,0.5) {x}; 
        \node at (1,0.2) {$z_0^{gt}$}; 
\end{tikzpicture}
\\  
\hline 
\end{tabular}

\caption{Positive canonical form}
\label{tab:constants}
\end{figure}

\noindent For Case A: $\displaystyle \left. 
        \dfrac{\partial a\big(i_{gt}(b))\big)}{ \partial z_{0}} \right. = 0$ and $\displaystyle \left. 
        \dfrac{\partial d_{gt}(b)}{ \partial z_{0}} \right. = -E_2$: 
    \begin{equation}
    \forall z_{0} \in \hspace{1mm}]-\infty, z_0^{gt}[, \dfrac{\partial IoU_{gt}(b)}{\partial z_{0}} = \dfrac{ E_0\cdot E_3 \cdot E_2}{d_{gt}(b)^2} \geq 0
    \label{maxcase1}
    \end{equation}

\noindent For Case B: $\displaystyle \left. 
        \dfrac{\partial a\big(i_{gt}(b))\big)}{ \partial z_{0}} \right. = -E_0$ and $\displaystyle \left. 
        \dfrac{\partial d_{gt}(b)}{ \partial z_{0}} \right. = (E_0-E_2)$:
     \begin{equation}
    \forall z_{0} \in \hspace{1mm}] z_0^{gt}, +\infty[,  \dfrac{\partial IoU_{gt}(b)}{\partial z_{0}} = \dfrac{-E_0 \cdot (E_2 \cdot E_5 + a(b_{gt}) )}{d_{gt}(b)^2} \leq 0
    \label{maxcase2}
    \end{equation} 

At fixed $z_{1-3}$ and given that \(IoU_{gt}(b)\) is increasing for \(z_0 < z_0^{gt}\) (equation \ref{maxcase1}), decreasing for \(z_0 > z_0^{gt}\) (equation \ref{maxcase2}) and being continuous at \(z_0 = z_0^{gt}\), \(IoU_{gt}\) reaches a local maximum at \(z_0 = z_0^{gt}\) regardless of the values of \(z_1\), \(z_2\), and \(z_3\). Similarly, for each $z_i$ coordinate, and fixing others constant, \(IoU_{gt}\) reaches local maximum at \(z_1 = z_1^{gt}\), \(z_2 = z_2^{gt}\), \(z_3 = z_3^{gt}\).

\end{document}